\documentclass[runningheads]{llncs}
\usepackage{graphicx}
\usepackage{comment}
\usepackage{amsmath,amssymb} 
\usepackage{color}
\usepackage{multirow}
\usepackage{float}
\usepackage{subfigure}
\usepackage{bm}
\usepackage{url}

\begin{document}
\pagestyle{headings}
\mainmatter
\def\ECCVSubNumber{458}  

\title{Asymmetric Rejection Loss for Fairer Face Recognition} 

\author{Haoyu Qin}
\institute{{hq2172@columbia.edu}}
\maketitle

\begin{abstract}
Face recognition performance has seen tremendous gain in the recent years, mostly due to the availability of large-scale face images dataset that can be exploited by deep neural networks to learn powerful face representations. 
However, recent research has shown differences of face recognition performance across different ethnic groups mostly due to the racial imbalance in the training datasets where Caucasian identities largely dominate other ethnicities. 
This is actually symptomatic of the under representation of non-Caucasian ethnic groups in the celebdom from which face datasets are usually gathered, rendering the acquisition of labeled data of the under-represented groups challenging.
In this paper, we propose an Asymmetric Rejection Loss, which aims at making full use of unlabeled images of those under-represented groups, to reduce the racial bias of face recognition models. 
We view each unlabeled image as a unique class, however as we cannot guarantee that two unlabeled samples are from distinct class we exploit 
labeled and unlabeled data in an asymmetric manner in our loss formalism.
Extensive experiments show our method's strength in mitigating racial bias, outperforming state-of-the-art semi-supervision methods. 
Performance on the under-represented ethnicity groups increases while that on well-represented group is nearly unchanged.

\end{abstract}


\section{Introduction}
Face recognition reached a high level of performance in recent years~\cite{Arcface,han2018face,cosface,R3}, along with the development of convolution neural network~\cite{resnet,shufflenetv2,mobilenetv2,efficientnet,legonet}. However, as pointed out by several works \cite{alvi2018turning,amini2019uncovering,RFW}, model bias is one urgent issue to be solved in this field. The main cause of the model bias between well-represented groups and under-represented groups is the distribution of training dataset. As shown in \cite{RFW}, we can easily observe that the commonly used face recognition datasets \cite{vggface2,ms1m,LFW,IJB-A,CASIA} are dominated by Caucasian identities. Since the dataset is mainly formed by Caucasian subjects, face recognition models’ performance on Caucasian outperforms that on other group of people, such as African, Asian, and Indian. Similarly, gender is another  aspect of face recognition datasets' imbalance. Dataset is mainly consist of male faces. Much attention are needed to pay to this field to make face recognition models be fairer.

The number of identities all over the world is fairly large. Therefore, introducing some new under-represented group identities into the training process to balance the training process seems like an easy and promising methodology. However, since annotating images for a large number of classes is usually prohibitive, fully-supervised training is not suitable. Thus, we try to utilize unlabeled images in this paper. It worth to mention that we cannot simply classify each unlabeled image into a class and use self-training approach as \cite{xie2019self} because the labeled classes is quite limited and unlabeled images are less likely to belong to this close set.

When unlabeled images are collected from complex sources, e.g. website, the number of identities is always much greater than the average images per subject. So, clustering is not suitable for this case because it's hard to obtain large clusters without noise. If the scale of clusters is small and lacks divergence, the clustering is meaningless. If clusters contain noise, as pointed by \cite{devil}, models' performance could be dampened a lot. Therefore, unlike Transductive Centroid Projection (TCP) \cite{liu2018transductive}, we don't take clustering into consideration and we leverage the unlabeled data in a similar way as \cite{UIR}.
\begin{figure}[!htbp]
\centering
\subfigure[Asymmetric Rejection Loss]{
\label{Fig.sub.1}
\includegraphics[width=0.45\textwidth]{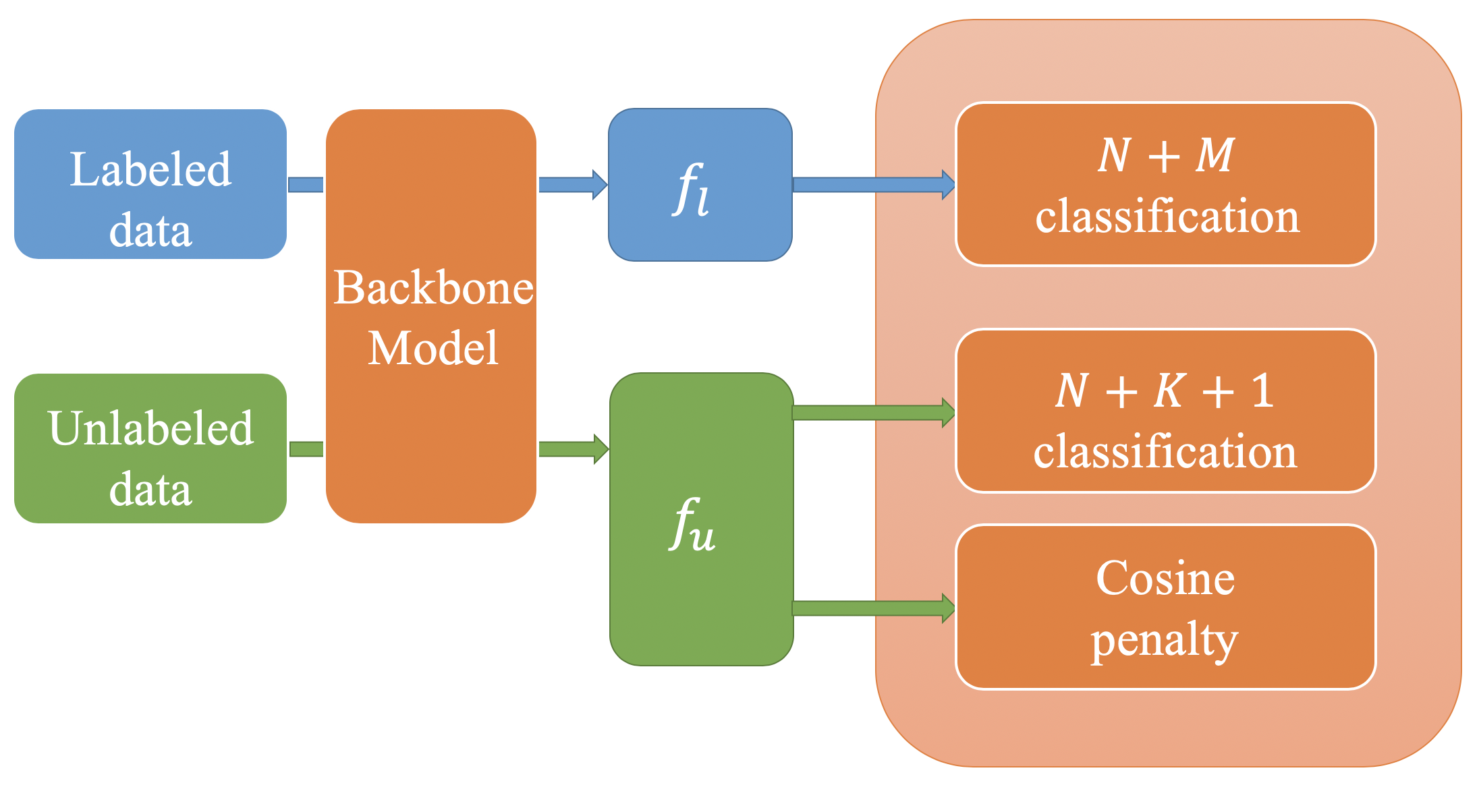}}
\subfigure[Unknown Identity Rejection Loss]{
\label{Fig.sub.2}
\includegraphics[width=0.45\textwidth]{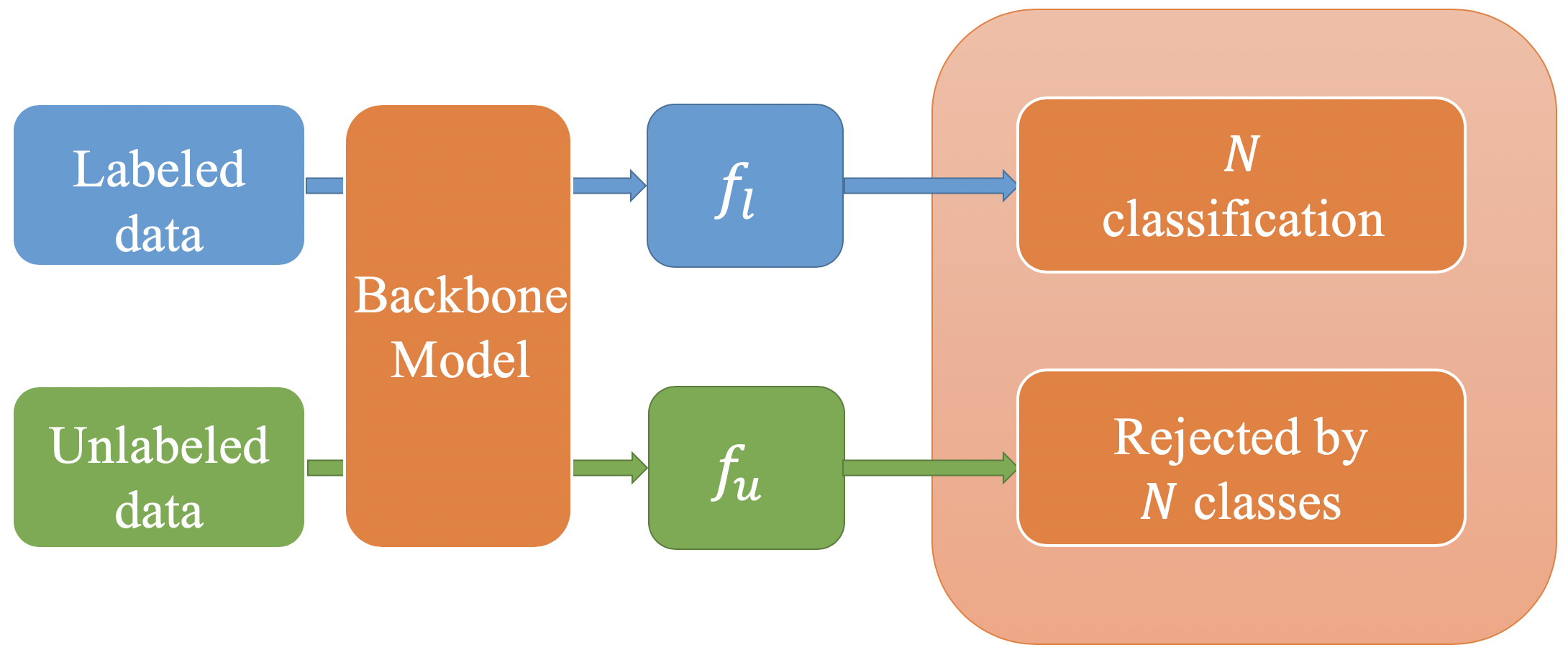}}
\caption{A pipeline comparison between (a) Asymmetric Rejection Loss with cosine penalty and (b) Unknown Identity Rejection Loss. Here are $N$ classes in labeled dataset and $M$ selected unlabeled images, which are not overlapped with $N$ classes. 1 represents the current input unlabeled image and $K$ stands for the number of images that have a different ethnicity label with current input unlabeled image.}
\label{pipeline}
\end{figure}

 In \cite{UIR}, the authors design an Unknown Identity Rejection (UIR) loss that requires unlabeled data being ``rejected'' by annotated classes in labeled training dataset, shown in Figure \ref{Fig.sub.2}. In this paper, we point out the weakness of their method, and modify it for the first step to alleviate aforementioned model racial bias. We propose a new semi-supervised method, Asymmetric Rejection Loss (ARL), as shown in Figure  \ref{Fig.sub.1}. 
 
 In this study, following TCP \cite{liu2018transductive} and UIR \cite{UIR}, we firstly filter out the possible overlap between labeled dataset and unlabeled images. By viewing each unlabeled image as a unique class, unlabeled data are utilized in a supervised way as \cite{liu2018transductive}, but in our approach unlabeled data and labeled data are treated asymmetrically. Specifically, for labeled images, not only all the $N$ labeled classes but also the entire $M$ unlabeled images are added into classification, while for unlabeled images, we add the current unlabeled image, $K$ unlabeled images that have a different ethnicity tag and $N$ labeled classes into classification. In a nutshell, because of the lack of identity information for unlabeled images, we only conduct classification within the group of classes that is safe and compatible. Even though there are some images that belong to the same individual in the unlabeled dataset, this method can successfully avoid conflicting.
 
 Furthermore, we extend the thought of ARL and make full use of unlabeled image for the next step. It makes the model fairer by optimization on cosine similarity of safe unlabeled image pairs in a mini-batch. 

We evaluate our method on RFW test dataset, provided by \cite{RFW}, and extensive experiments on different commonly used face recognition training dataset show that our method truly mitigates the model bias on different racial groups. The contributions of this paper can be briefly summarized as:

a) We propose a novel perspective on leveraging unlabeled images and put forward a semi-supervised method.

b) Our method can increase the model performance on face verification task and greatly alleviate the model racial bias problem.

c) We further discuss the usage of unlabeled data and gender bias issue based on our framework.

\section{Related Works}
\subsection{Deep Face Recognition}
Novel loss functions are the focus in the field of face recognition in recent years \cite{Arcface,cosface,liu2017sphereface,triplet_loss,contrastive_loss,center_loss}. A bunch of loss functions are proposed to optimize the distance metrics in the feature space. Arcface \cite{Arcface}, which is based on the angular distance, namely cosine similarity, is one of the most outstanding loss functions. It takes both intra-class distance and inter-classes distance into account by introducing an additive angular margin to the loss function. Our proposed method, Asymmetric Rejection Loss, is based on Arcface \cite{Arcface}.

\subsection{Racial Bias in Datasets}

Many different natural attributes can be annotated to each individual in this world, such as gender, race, and age. Even though commonly used face recognition traing datasets, namely CASIA-Webface \cite{CASIA}, VGGFace2 \cite{vggface2}, and MS-Celeb-1M \cite{ms1m}, are large enough, containing a lot of subjects and images, they are still not balanced from the perspective of these attributes. Following \cite{RFW}, we mainly focus on the racial unbalance in this paper. 

As shown in Table \ref{racial_distribution}, except RFW \cite{RFW}, other datasets, no matter training dataset or testing dataset, are severely unbalanced on racial distribution. Because of the large number of subjects and images, RFW \cite{RFW} is a good dataset to evaluate face recognition model on different ethnicity group.

\begin{table}[!htbp]
\centering
\caption{The percentage of different race in commonly-used training and testing databases. Reported by RFW \cite{RFW}.}
\begin{tabular}{c|c|c|c|cccc}
\hline
\multirow{2}{*}{\begin{tabular}[c]{@{}c@{}}Train/ \\ Test\end{tabular}} & \multirow{2}{*}{Dataset} & \multirow{2}{*}{Subjects} & \multirow{2}{*}{Images} & \multicolumn{4}{c}{Racial Distribution (\%)} \\
 &  & & & African & Asian & Caucasian & Indian \\ \hline
\multirow{3}{*}{Train} & CASIA-Webface \cite{CASIA} & 10K & 0.5M & 11.3 & 2.6 & 84.5 & 1.6 \\
 & VGGFace2 \cite{vggface2} & 8.6K & 3.1M & 15.8 & 6.0 & 74.2 & 4.0 \\
 & MS-Celeb-1M \cite{ms1m} & 90K & 5.0M & 14.5 & 6.6 & 76.3 & 2.6 \\
\hline
\multirow{3}{*}{Test} & LFW \cite{LFW} & 5.7K & 13K & 14.0 & 13.2 & 69.9 & 2.9 \\
 & IJB-A \cite{IJB-A} & 0.5K & 5.7K & 17.0 & 9.8 & 66.0 & 7.2 \\
 & RFW \cite{RFW} & 12K & 40K & 25.0 & 25.0 & 25.0 & 25.0 \\ \hline
\end{tabular}
\label{racial_distribution}
\end{table}

\subsection{Semi-supervision for Face Recognition} How to leverage unlabeled data to promote model performance is a goal of semi-supervised learning \cite{chapelle2009semi,zhu2005semi}. For face recognition, some works \cite{gao2017semi,zhao2011semi} based on the assumption that the set of categories are shared between labeled data and unlabeled data, but it is not practical as we mentioned before. There are two novel works \cite{zhan2018consensus,liu2018transductive} based on clustering in this field. \cite{zhan2018consensus} is complicated since it builds up a committee, while \cite{liu2018transductive} is relative easy to implement.

\subsection{Transductive Centroid Projection}
 Transductive Centroid Projection (TCP) \cite{liu2018transductive} is a cluster based semi-supervision method. The main idea for this method is to use unlabeled data in a labeled manner. Authors in \cite{liu2018transductive} first cluster unlabeled images and give them pseudo-label. Then in each mini-batch, for unlabeled images with pseudo-label, authors select $l$ classes with $o$ images each class. Suppose there are $N$ classes in labeled dataset, TCP \cite{liu2018transductive} conducts classification on $N+l$ classes in each mini-batch.
 
 The idea of their work is inspiring and our Asymmetric Rejection Loss holds similar insight. Nevertheless, their method relies on the easy-to-cluster property of unlabeled images. So, the source of unlabeled images can be a limitation for their method.

\subsection{Unknown Identity Rejection Loss}
 Unknown Identity Rejection Loss \cite{UIR} tries to minimize the negative entropy for the classification probability distribution of unlabeled images. Their loss is rewritten in Equation \ref{eq1}. 

\begin{equation}
L = -\sum_{i=1}^{N} \log (p_i) = -\sum_{i=1}^{N} \log \frac{a_i}{\sum_{j} a_j}
\label{eq1}
\end{equation}
where $N$ is the number of classes in labeled dataset, $p_i$ is the probability on class $i$ and $a_i$ is the activation value of last FC-Layer on class $i$. 

After pre-processing, 
there is no overlap between labeled dataset and unlabeled images. So, their work is aiming at answering the question what's the probability 
distribution should be for an image of unseen identity on the labeled 
training dataset. Their answer is discrete uniform distribution, which 
means probability on all the labeled classes, namely $\frac{1}{N}$.

Intuitively, the lower $p_i$, the stronger distinguishing ability for the $i_{th}$ class and current unlabeled image. If we discard the constraint that $\sum_{i=1}^N p_i = 1$, the optimization goal should be $p_i = 0$ for every class in labeled trainings dataset. Therefore, UIR loss's optimization goal \cite{UIR} has some flaw. Besides, minimizing negative entropy is used to extract identity-irrelevant features in \cite{liu2018exploring}, which contradicts the goal of face recognition, extracting identity-related features. Since the sum of all possibilities should equal to one, inserting a new class into the class set should be a suitable solution. In this way, the optimization goal requires classification probability for all labeled classes to be zero.

\section{Approach}
In this section, we illustrate and formulate our method in detail. Since we point out the weakness of UIR \cite{UIR}, we first modify it by replacing entropy minimizing with two asymmetric classification tasks, which has more reasonable optimization goal. Besides, maintaining of unlabeled images' weight vectors in feature space enable all the unlabeled images, rather than selected $l$ classes, to be rejected in one mini-batch. Moreover, we extend the thought of rejection as the second part, asking the current unlabeled image to reject some other safe unlabeled data in a mini-batch. Here, the rejection is achieved by direct optimization on cosine similarity. Unlabeled images can be exploited better in this way. Finally, we explain how our method works from the geometric perspective.

\subsection{Asymmetric Rejection Loss}
In this study, we propose an asymmetric loss to conduct different tasks on labeled data and unlabeled data. Note that even though unlabeled data here don't have an identity label, they do have an ethnicity label. In addition, even though we treat labeled data and unlabeled images in different ways, as shown in Figure \ref{pipeline}, the weight of fully-connected layer is shared by them. First, following \cite{UIR} we eliminate all the possibly overlapped data according to the classification probability to make sure that there won't be images belonging to labeled identities in unlabeled dataset. Specifically, we classify unlabeled images with pretrained backbone model and discard those images that have a classification probability greater than $0.9$. Then each unlabeled image is viewed as a unique class, which means it has its own weight vector in the fully-connected layer.

We therefore propose to actually define two different problems, depending on whether the input sample is a labeled or unlabeled sample in a random batch of size $B$.

In the case where the input sample is an unlabeled sample, we consider all the weights of the $N$ known classes plus a weight vector corresponding to that single unlabeled sample, initialized with its feature. Furthermore, we can introduce all the images that have a different ethnicity label into the classification. Now, this unlabeled data classification task is a $N+K+1$ classification, where $N$ is the number of classes in labeled dataset and $K$ is number of classes in unlabeled dataset that has a different ethnicity label. In that setting, the model should maximize the probability corresponding to the weight vector of that unlabeled sample while diminishing as much as possible the probability of all known classes. In this way, we don't need to consider any other unlabeled sample and determine whether or not they are from the same identity.

In the case where the input sample is a labeled sample, we still consider all the weights of the $N$ known classes plus all the $M$ weight vectors corresponding to all the unlabeled samples in the current batch. In that setting, the model should maximize the probability corresponding to the correct class of that labeled sample while diminishing as much as possible the probability of all other classes including the probability of the $M$ unlabeled samples.

We implement this strategy using the ArcFace\cite{Arcface} loss which relies on a modification of the standard Softmax computation by adding an angular margin $m$ to the angle $\theta_{y_i}$ between a sample and its target class weight. Formally, denoting $s$ as the scale parameter, the ArcFace \cite{Arcface} loss for the $B_U$ unlabeled samples in a batch is computed as Equation \ref{eq2}.

\begin{equation}
L_U = - \frac{1}{B_U} \sum_{i=1}^{B_U} \log \frac{e^{s cos(\theta_{y_i}+m)}}{e^{s cos(\theta_{y_i}+m)} + \sum_{j=1, j\neq y_i}^{N+K+1} e^{s cos \theta_j} }
\label{eq2}
\end{equation}
where $\theta_{y_i}$ is the angle between the $i^{th}$ unlabeled sample in the batch and its own weight vector. 

While the loss for $B-B_U$ labeled samples can be formulated as Equation \ref{eq3}.

\begin{equation}
L_L = - \frac{1}{B-B_U} \sum_{i=1}^{B-B_U} \log \frac{e^{s cos(\theta_{y_i}+m)}}{e^{s cos(\theta_{y_i}+m)} + \sum_{j=1, j\neq y_i}^{N+M} e^{s cos \theta_j}}
\label{eq3}
\end{equation}
where $\theta_{y_i}$ is the angle between the $i^{th}$ labeled sample in the batch and the weight vector of that class. $N+M$ classification on labeled data maintains the wellness of labeled classes' weight vector, while $N+K+1$ classification keeps the meaning of unlabeled images' weight vector. These two classification tasks are interdependent.

\subsection{Rejection between Safe Unlabeled Image Pairs}

\begin{figure}[!htbp]
\centering
\subfigure[African]{
\label{African}
\includegraphics[width=0.31\textwidth]{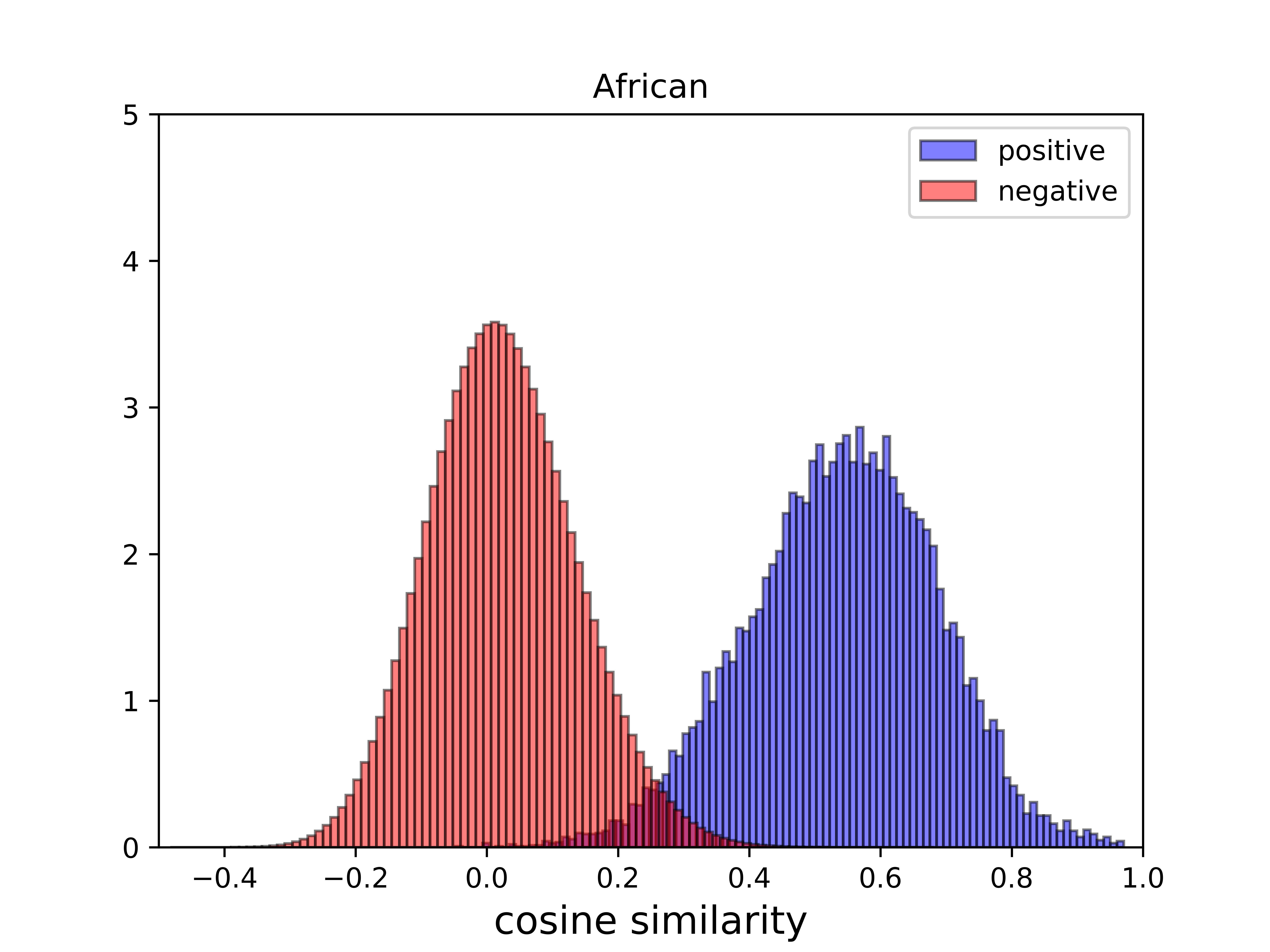}}
\subfigure[Asian]{
\label{Asian}
\includegraphics[width=0.31\textwidth]{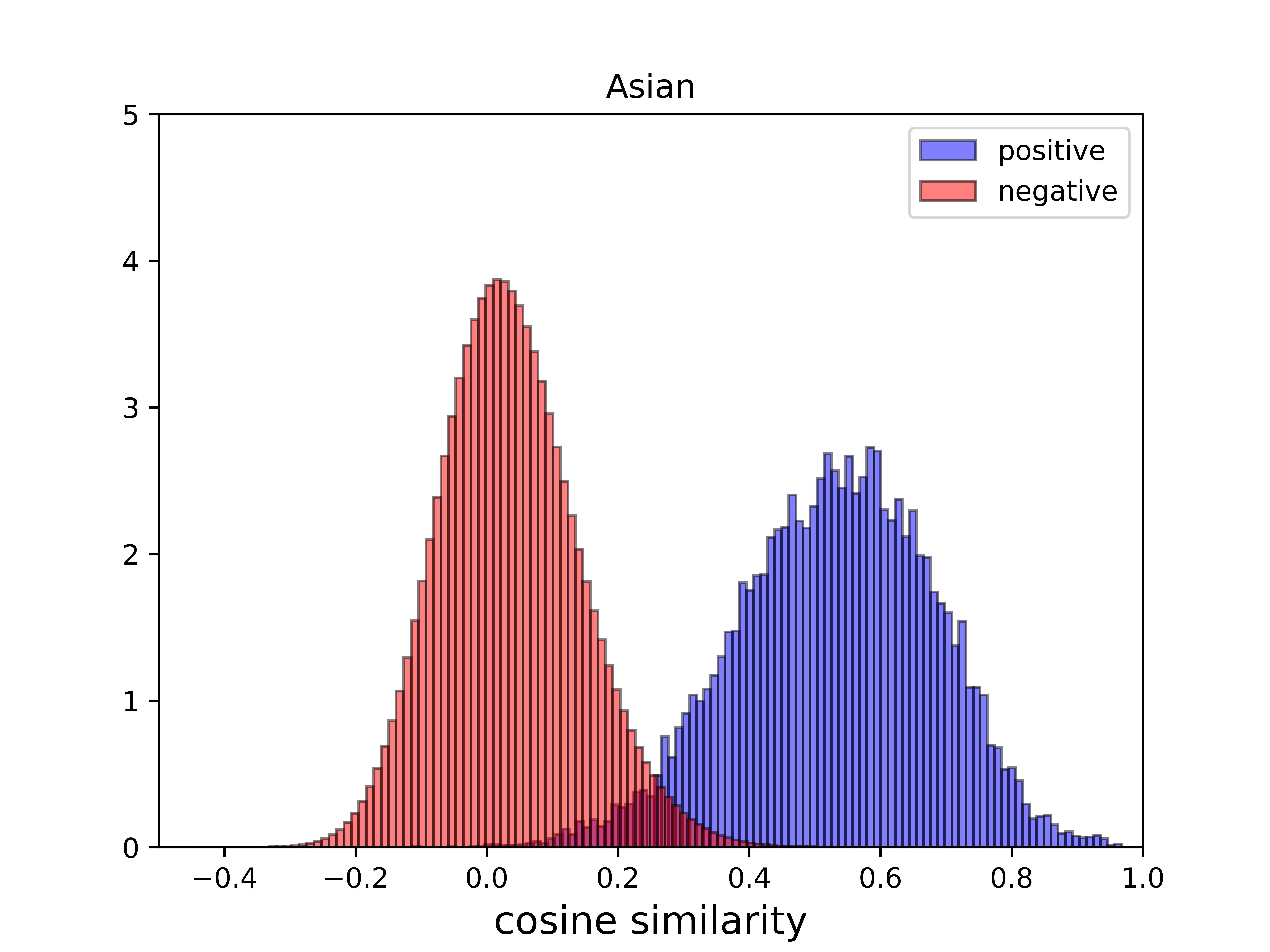}}
\subfigure[Indian]{
\label{Indian}
\includegraphics[width=0.31\textwidth]{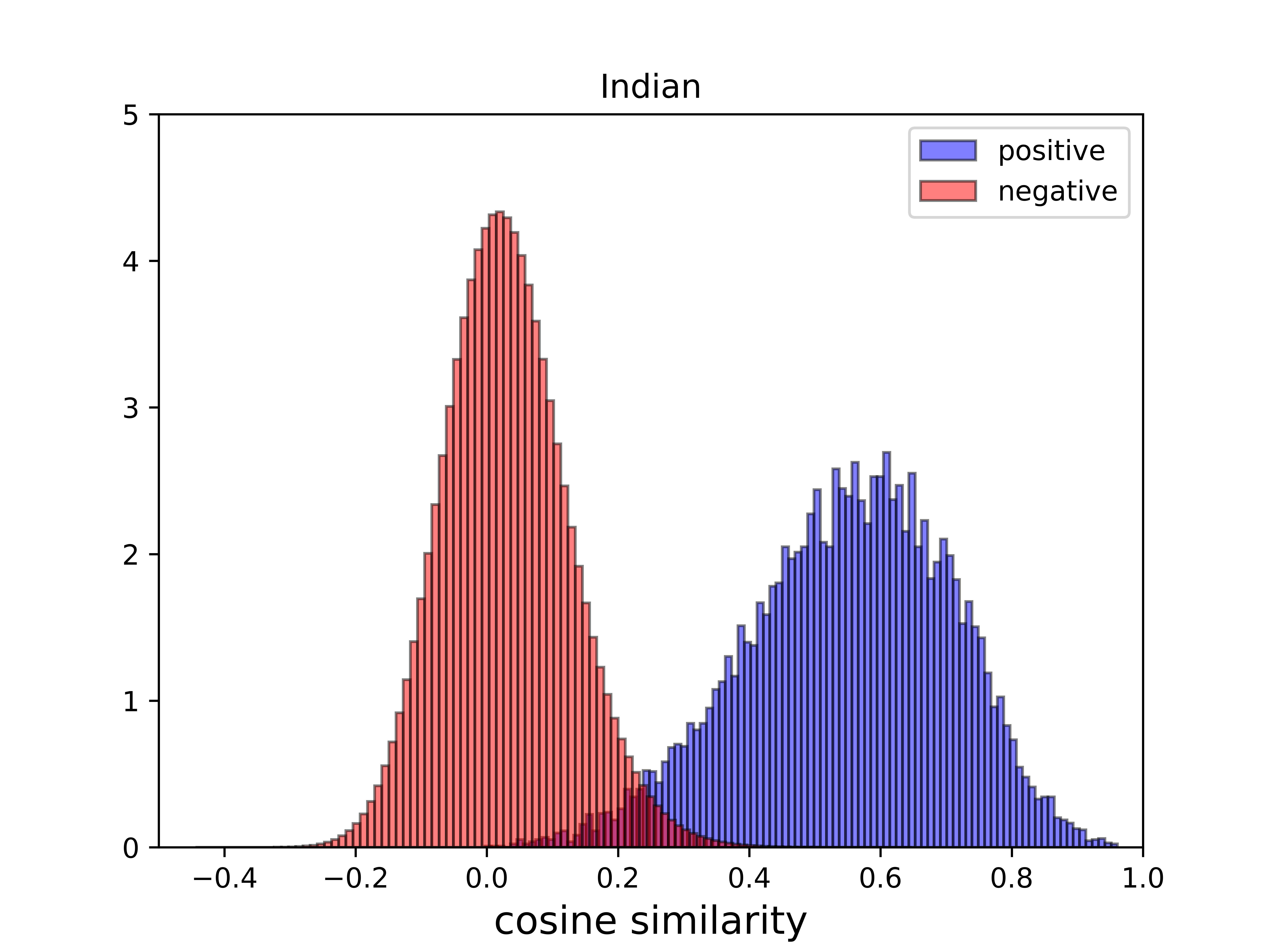}}
\caption{Normalized cosine similarity histograms of positive pairs and negative pairs on each under-represented group. Blue represents positive pairs, while red represents negative pairs.}
\label{Hist}
\end{figure}

Normally, the number of identities is greatly larger than the average number of images per subject when unlabeled images are collected from website. So, when we grab a batch of unlabeled images from dataset, it is likely that all pairs of images don't belong to the same identity. Thus, it is feasible and profitable to distinguish such pairs. So, we propose a cosine penalty strategy to use one samples to ``reject'' all the other unlabeled samples that would not belong to the same subject in a mini-batch. In order to avoid introducing possible conflict, we can add some constraint to the selection of negative pairs in a mini-batch. Since our baseline model is valid, the angular distance between features of the same identity are relatively small. As shown in Figure \ref{Hist}, a valid model has separable normalized cosine similarity histograms for positive pairs and negative pairs even on under-represented ethnicity groups.

Therefore, we can easily select a threshold, noted as $t$, at which the true positive rate is fairly large and the false positive rate is small enough. Thus, we can assume that pairs of unlabeled images with cosine similarity lower than $t$ don't belong to the same identity and safely optimize their cosine distance.

Moreover, to boost the performance, our model should focus on those features that haven't been well learned. So, we also set a lower bound for the cosine penalty strategy. Usually, researchers assume that when the feature vectors are orthogonal to each other, they are well represented, so we don't further add penalty on negative pairs that have a cosine similarity lower than zero. 

Up to now, we establish a cosine similarity interval, $(0, t)$. We assume that any pair of faces with a cosine similarity locates in this interval is a valuable negative pair and we directly optimize on the cosine similarity. As for the specific loss function, we choose L2 loss that measure the mean squared error. Detailed loss function can be formulated as Equation \ref{eq4}.

\begin{equation}
L_C = \frac{\sum_{i, j}(cos(f_i, f_j)^2)}{N_t}, 0 < cos(f_i, f_j) < t
\label{eq4}
\end{equation}
where $f_i$ and $f_j$ are the normalized feature vector for unlabeled image $i$ and $j$ , $t$ is the upper bound of penalty interval, and $N_t$ is the number of feature pairs whose cosine similarity locates in the interval $(0, t)$.

As many other works, we combine those two parts of loss function linearly. Thus, the whole loss function can be written in the form of Equation \ref{eq5}.

\begin{equation}
L = L_L + \lambda_U \cdot L_U + \lambda_C \cdot L_C
\label{eq5}
\end{equation}
where $\lambda_U$ and $\lambda_C$ are two loss weights.

\subsection{Geometrical Interpretation}

In this part, we try to give out an explanation with geometric view. Novel face recognition loss functions, such as Arcface \cite{Arcface} and CosFace \cite{cosface}, are designed to pursue a feature space with good distribution. The distribution should be compact regarding to feature vectors from one class, which the distance between feature vectors from different identities should be sufficiently far.

\begin{figure}[!htbp]
\centering
\subfigure[]{
\label{Ori_feature_space}
\includegraphics[width=0.38\textwidth]{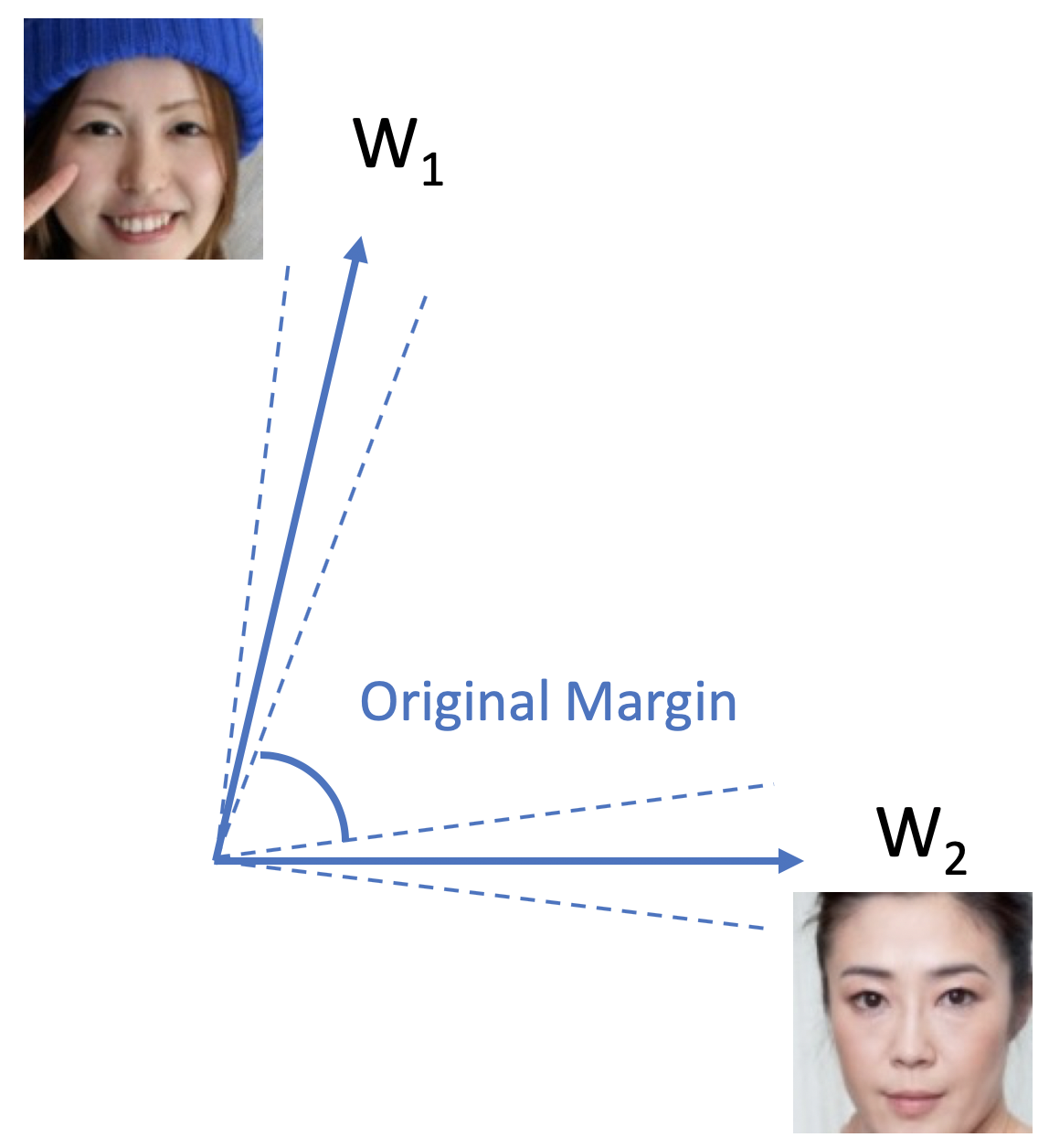}}
\subfigure[]{
\label{Feature_space}
\includegraphics[width=0.52\textwidth]{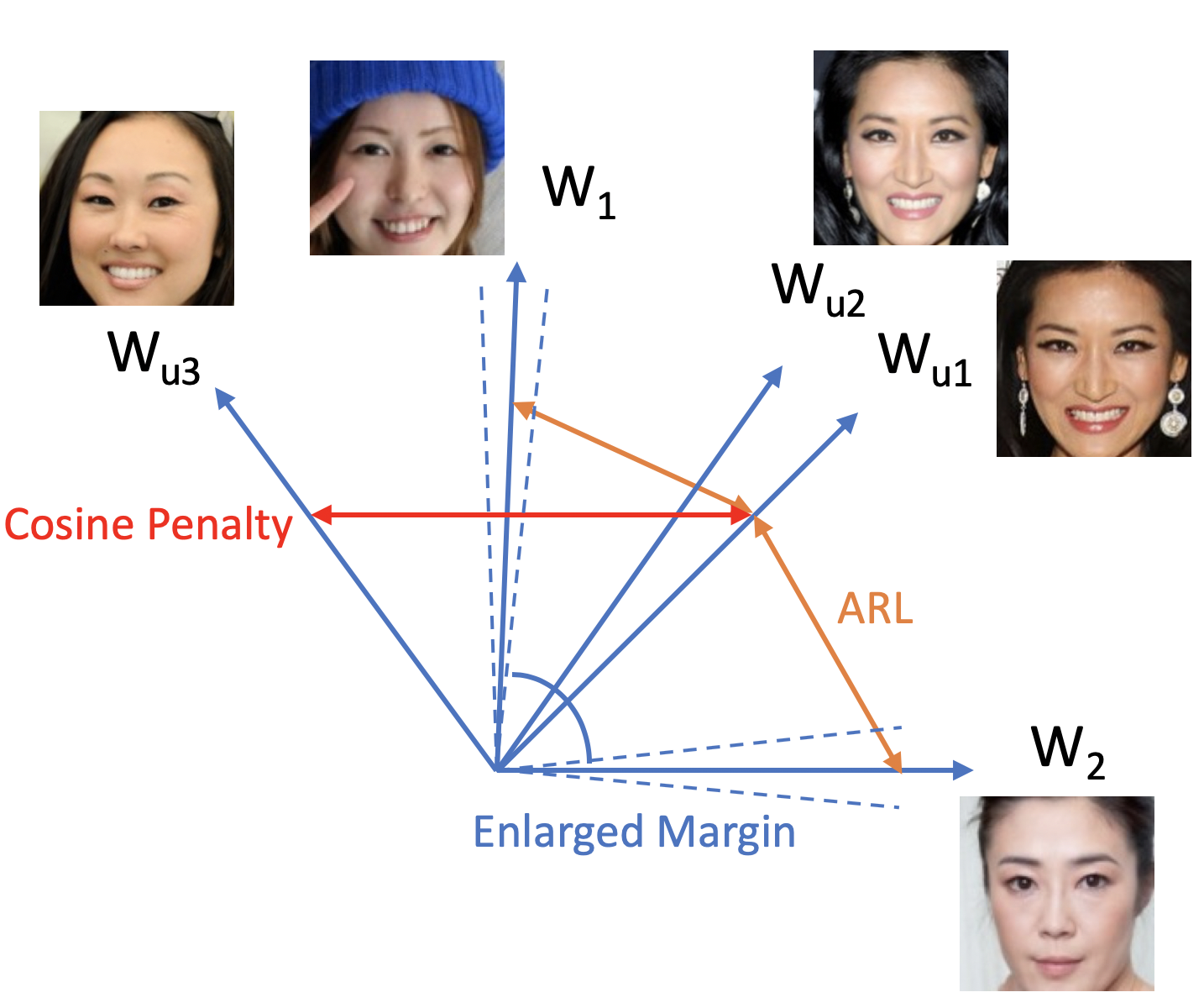}}
\caption{Left sub-figure illustrates the scenario of baseline model, and the right sub-figure illustrates our proposed method. The dash lines represent the decision boundary of labeled classes. $W_1$ and $W_2$ are weight vectors for two labeled classes. $W_{u_1}$, $W_{u_2}$, and $W_{u_3}$ are weight vectors for three unlabeled images. In the right sub-figure, all two-way arrows illustrate the loss items that are related to $W_{u_1}$. Since there are unlabeled images belonging to the same person, like $W_{u_1}$ and $W_{u_2}$, unlabeled images that have the same ethnicity label will not take part in $N+K+1$ classification. Cosine penalty will only be added on negative pairs such as $W_{u_1}$ and $W_{u_3}$ that the distances are large enough to avoid conflicts.}
\label{feature_space_compare}
\end{figure}

Our method is based on the assumption that weight vectors represent the class well. Then inserting unknown identities into the feature space benefits the feature distribution, as demonstrated in Figure \ref{feature_space_compare}. A newly inserted weight vector makes the decision margin become insufficient because this new class has to be far enough from labeled classes. By those two classification tasks, this new class will hold some space and make some labeled classes' feature vectors be more compact. Thus, ARL can enlarge the distance between labeled classes, which means being more distinguishable. Cosine penalty strategy are adopted to enlarge distance among unlabeled images for the same reason. According to independently identically distribution assumption, the noramlized cosine similarity histograms for unseen data should be more separable.

\section{Experiments}
\subsection{Experiment Settings}
The backbone model of our experiments is the modified ResNet-34 architecture mentioned in \cite{Arcface}. And Arcface Loss \cite{Arcface} is adopted for all the experiments. Five facial landmarks are used for face alignment, then cropped images are resized to the resolution of $112\times112$. Besides, each pixel of input images is normalized to the range of $[0, 1]$. We train all the experiments on 2 GPUs with a batch size of 256. SGD optimizer is adopted and momentum is set to be 0.9. Weight decay is $5e-4$ in our experiments. 

Following \cite{UIR}, we finetune the baseline model with setting the ratio, between labeled images and unlabeled images in a mini-batch, as 3:1. As many works based on Arcface \cite{Arcface}, we set the scale factor $s$ as 64 and the additive margin $m$ as 0.5. The loss weight for labeled images is 1, while for unlabeled images is 3. For cosine penalty, we set the threshold as 0.3 and set the loss weight to be 10.

\subsection{Datasets}
\subsubsection{Evaluation Dataset}
We evaluate our model on RFW test dataset \cite{RFW} since the large number of identities and images shown in Table \ref{racial_distribution} and the reliable manually given ethnicity label. All possible pairs within the same ethnicity group are considered. As reported in \cite{RFW}, in one ethnicity group, there are about 14K positive pairs and 50M negative pairs. We find that the cross-group negative pairs are well learned by face recognition model, since faces from different ethnicity truly look more different. So, we focus on the model's performance within the same ethnicity.

Note that RFW test dataset \cite{RFW} is balanced from the view of ethnicity but unbalanced on gender aspect. From the figure in \cite{RFW}, we can find out that male faces is much more than female faces, especially for group African, 95 percent of whose images are male. The male percentage of group Caucasian and group Indian is about 70 percent, while that of group Asian is lower than 60. Such an imbalance has effect on the evaluation results, and we discuss this gender issue in this paper.

\subsubsection{Labeled Training Dataset}
CASIA-Webface \cite{CASIA}, VGGFace2 \cite{vggface2}, and MS-Celeb-1M \cite{ms1m} are adopted as labeled training dataset in our experiments. And note that since RFW test dataset is selected from MS-Celeb-1M \cite{ms1m}, the overlapping part of dataset is eliminated from training process. In CASIA-Webface \cite{CASIA}, there are 10572 identities and 490,623 images. In VGGFace2 \cite{vggface2}, there are 8631 identities and 3,137,807 images. After eliminating overlap identities, MS-Celeb-1M \cite{ms1m} remains 3,459,591 images that belong to 75460 subjects, and we denote this dataset as MS\_wo\_RFW \cite{RFW} below.

As far as we known, VGGFace2 \cite{vggface2} is a nearly gender balanced dataset, with over 40 percent of female. We use a tool \cite{attr} to estimate the ratio for CASIA-Webface \cite{CASIA} and MS\_wo\_RFW \cite{RFW}. The former one is balanced, while the latter one is biased. The Male-Female ratio for MS\_wo\_RFW \cite{RFW} is about 2:1.

\subsubsection{Unlabeled Training Dataset}
As for the unlabeled dataset, we use the unlabeled part of RFW training dataset \cite{RFW}, namely African, Asian, and Indian subsets. In order to make sure that there is no overlap between labeled dataset and unlabeled dataset, we use the strategy mentioned in \cite{UIR}. First, unlabeled images are classified on labeled dataset. Then, we discard images with a probability over 0.9. In this way, we assume that the intersection of labeled and unlabeled data is empty.

Remaining images are viewed as valid images that can be used in training process. We sort the images according to the magnitude of their feature vectors and select first 20 thousand images in each ethnicity. It is because that images may contain more id-irrelevant features when the magnitude of feature vector is small. Since our method teaches the model in the form of mutual rejection without intra-class clustering, id-irrelevant features can do harm to the training process. 

\subsection{Performance Evaluation}
\subsubsection{Racial Bias}

\begin{figure}[!htbp]
\centering
\subfigure[CASIA-Webface]{
\label{casia}
\includegraphics[width=0.31\textwidth]{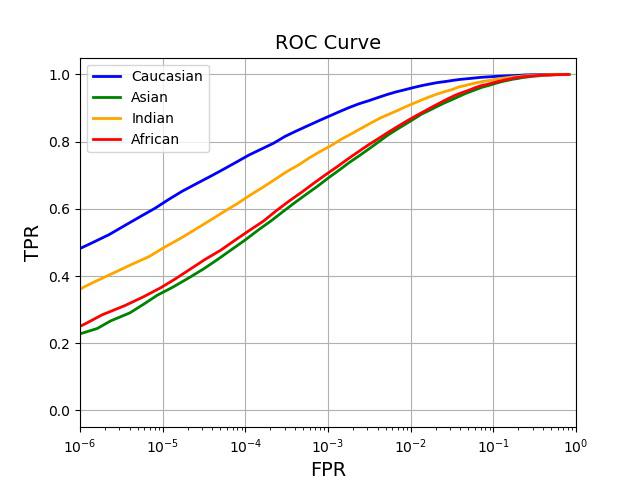}}
\subfigure[VGGFace2]{
\label{vgg}
\includegraphics[width=0.31\textwidth]{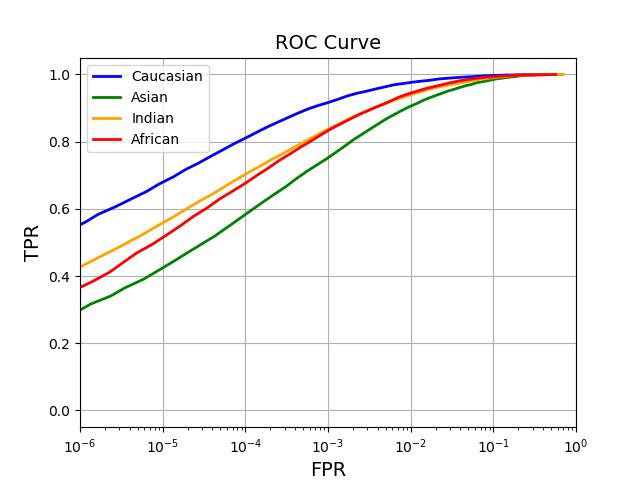}}
\subfigure[MS-Celeb-1M]{
\label{ms1m}
\includegraphics[width=0.31\textwidth]{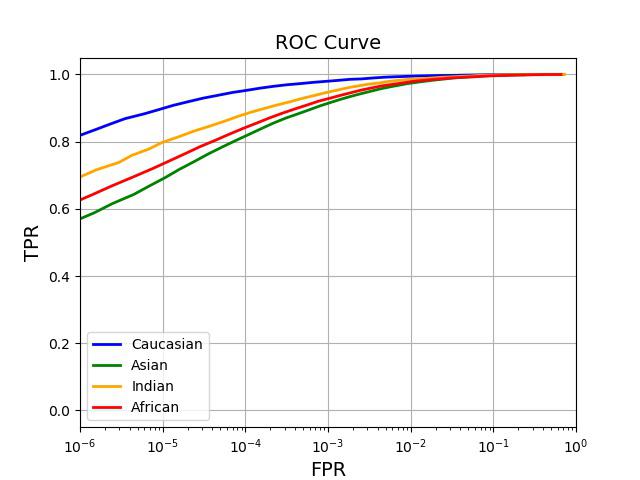}}
\caption{Racial bias can be observed from the ROC of baseline models. Performance on group Caucasian outperforms that on other under-represented groups significantly.}
\label{ROC}
\end{figure}

We trained baseline models as \cite{RFW}. From Figure \ref{ROC}, the ROC of three baseline models, it is obvious that face recognition models trained on commonly used face recognition dataset have significant racial bias. Our result is quite similar to the curves reported in \cite{RFW}. We can observe that performance on group Caucasian (blue curve) is much better than the other three ethnicity groups, especially group Asian (green curve) and group African (red curve).

\subsubsection{ARL Mitigates Racial Bias}
We train face recognition models with three labeled dataset mentioned above to evaluate our Asymmetric Rejection Loss. Based on the result shown in Table \ref{CASIA}, \ref{VGGFace2}, and \ref{MS1M}, it is obvious that our method does have the capability to alleviate the racial bias in face recognition models. In these three tables, we can observe the performance on under-represented groups increases a lot, while the performance on group Caucasian doesn't change too much. 

Besides, when comparing with UIR \cite{UIR}, we find that at most points, especially normal working point (FPR=$1e-4$), our method's results is higher than that of UIR \cite{UIR}. It is because that ARL has a more reasonable optimization goal than UIR \cite{UIR}. Besides, we maintain unlabeled images' weight vectors in feature space, which is able to exploit unlabeled data better since the existing of unlabeled images' weight vectors allows all the unlabeled images to participate in the loss calculation in each mini-batch.

Furthermore, we compare ARL with TCP \cite{liu2018transductive} on MS\_wo\_RFW \cite{RFW}. Since their method is based on clustering, a good baseline model is essential for successful clustering. So we use the baseline model trained on MS\_wo\_RFW \cite{RFW}. Nevertheless, their improvement is not as large as ARL. It is becasue that RFW training dataset \cite{RFW} does not possess good cluster property and we can not grab a great number of images in one cluster ($o$ in TCP \cite{liu2018transductive}) in each mini-batch. However, In TCP \cite{liu2018transductive}, the face images are collected from surveillance videos, which possess the easy-to-cluster property. They can easily obtain clusters with an average image number of 22. However, for RFW training dataset \cite{RFW}, it's rare to have a cluster with more than 5 images. For group Asian, the performance even drops. It is because that the baseline model has the worst performance on group Asian, so the clustering result for group Asian may contain too much noise, which leads to the performance decrements. Since ARL doesn't require unlabeled data to have the easy-to-cluster property, our method is more suitable in this case.

\begin{table}[!htbp]
\centering
\caption{Experiments on CASIA-Webface \cite{CASIA}. ``+C'' represents applying cosine penalty strategy, and ``+G'' means taking gender in to account for image selection.}
\scalebox{0.8}{
\begin{tabular}{c|c|cccccc}
\hline
\multirow{2}{*}{Ethnicity}& \multirow{2}{*}{Model}& \multicolumn{6}{c}{TPR@FPR} \\ \cline{3-8}
& & 1e-1 & 1e-2 & 1e-3 & 1e-4 & 1e-5 & 1e-6 \\ \hline
 \multirow{5}{*}{African} 
& baseline &  97.52&  86.74&  70.58&  52.62&  36.79&  24.86 \\
& UIR &  97.62(0.10)&  86.77(0.03)&  70.62(0.04)&  53.00(0.38)&  37.26(0.47)&  24.00(-0.86) \\
& ARL &  97.97(0.45)&  88.70(1.96)&  73.56(2.98)&  56.70(4.08)&  \bm{$41.45(4.66)$}&  \bm{$29.04(4.18)$} \\
& ARL+C &98.28(0.76)&  \bm{$89.51(2.77)$}&  \bm{$74.47(3.89)$}&  \bm{$57.80(5.18)$}&  41.31(4.52)&  28.11(3.25) \\
& ARL+C+G &  \bm{$98.32(0.80)$}&  89.40(2.66)&  74.17(3.59)&  57.11(4.49)&  41.03(4.24)&  27.91(3.05) \\ \hline
 \multirow{5}{*}{Asian} 
& baseline & 97.14&  86.11&  69.08&  50.74&  35.00&  22.67 \\
& UIR & 97.38(0.33)&  86.33(0.22)&  67.59(-1.49)&  47.57(-3.17)&  30.62(-4.38)&  17.77(-4.90) \\
& ARL &  97.38(0.24)&  87.31(1.20)&  71.40(2.32)&  53.94(3.20)&  38.09(3.09)&  26.30(3.63) \\
& ARL+C &\bm{$98.03(0.89)$}&  \bm{$89.29(3.18)$}&  \bm{$73.64(4.56)$}&  55.50(4.76)&  \bm{$38.53(3.53)$}&  \bm{$26.74(4.07)$} \\
& ARL+C+G&97.85(0.71)&  88.84(2.73)& 73.25(4.17)&  \bm{$55.53(4.79)$}&  \bm{$38.53(3.53)$}&  26.25(3.58) \\
\hline
 \multirow{5}{*}{Caucasian} 
& baseline & 99.36&  95.86&  87.40&  75.32&  61.51&  48.00 \\
& UIR & 99.31(-0.05)&  95.83(-0.03)&  87.55(0.15)&  75.52(0.20)&  61.69(0.18)&  48.65(0.65)\\
& ARL & 99.46(0.10)&  96.08(0.22)&  87.90(0.67)&  76.86(1.54)&  64.24(2.73)&  51.50(3.50) \\
& ARL+C & 99.39(0.03)&  96.07(0.21)&  87.90(0.50)&  76.24(0.92)&  63.16(1.65)&  49.28(1.28) \\
& ARL+C+G & 99.41(0.05)&  95.87(0.01)&  87.87(0.47)&  76.36(1.04)&  63.06(1.55)&  50.27(2.77) \\   \hline
 \multirow{5}{*}{Indian} 
& baseline & 98.42&  90.99&  78.27&  63.00&  48.15&  35.96 \\
& UIR & 98.66(0.24)&  91.89(0.90)&  79.47(1.20)&  64.50(1.50)&  50.14(1.99)&  37.65(1.69) \\
& ARL & 98.85(0.43)&  93.14(2.15)&  83.06(4.79)&  70.14(7.14)&  57.05(8.90)&  43.60(7.64) \\
& ARL+C & 98.92(0.50)&  93.61(2.62)&  \bm{$84.04(5.77)$}&  \bm{$71.61(8.61)$}&  \bm{$58.63(10.48)$}&  \bm{$45.60(9.64)$} \\
& ARL+C+G & \bm{$99.00(0.58)$}&  \bm{$93.73(2.74)$}& 83.93(5.66)&  71.19(8.19)&  58.29(10.14)&  45.22(9.26) \\ \hline
\end{tabular}
\label{CASIA}
}
\end{table}

\begin{table}[!htbp]
\centering
\caption{Experiments on VGGFace2 \cite{vggface2}. ``+C'' represents applying cosine penalty strategy, and ``+G'' means taking gender in to account for image selection.}
\scalebox{0.8}{
\begin{tabular}{c|c|cccccc}
\hline
\multirow{2}{*}{Ethnicity}& \multirow{2}{*}{Model}& \multicolumn{6}{c}{TPR@FPR} \\ \cline{3-8}
& & 1e-1 & 1e-2 & 1e-3 & 1e-4 & 1e-5 & 1e-6 \\ \hline
 \multirow{5}{*}{African} 
& baseline &  99.27&  94.40&  83.23&  67.60&  51.32&  36.48 \\
& UIR &  99.34(0.07)&  94.28(-0.12)&  83.34(0.11)&  67.45(-0.15)&  51.62(0.30)&  36.16(-0.32) \\
& ARL &  99.26(-0.01)&  94.26(-0.14)&  83.06(-0.17)&  68.12(0.52)&  52.42(1.10)& 36.97(0.49) \\
& ARL+C &  99.31(0.04)&  94.56(0.16)&  83.19(-0.04)&  68.40(0.80)&  \bm{$52.61(1.29)$}& \bm{$38.54(2.06)$}\\
& ARL+C+G &  \bm{$99.42(0.15)$}& \bm{$ 94.58(0.18)$}&  \bm{$83.89(0.66)$}&  \bm{$68.74(1.14)$}&  52.39(1.07)&  36.86(0.38) \\\hline
 \multirow{5}{*}{Asian} 
& baseline & 98.48&  90.54&  75.14&  58.20&  42.32&  29.74 \\
& UIR &  98.72(0.24)&  91.91(1.37)&  77.89(2.75)&  60.99(2.79)&  44.93(2.61)&  32.12(2.38) \\
& ARL &  98.61(0.13)&  92.13(1.59)&  78.95(3.81)&  62.46(4.26)&  46.32(4.00)&  32.54(2.80) \\
& ARL+C & \bm{$98.94(0.46)$}&  \bm{$93.20(2.66)$}&  \bm{$80.30(5.15)$}&  63.37(5.17)&  \bm{$46.75(4.43)$}&  \bm{$32.62(2.88)$} \\
& ARL+C+G &  98.84(0.36)&  92.90(2.36)&  79.95(4.81)& \bm{$63.48(5.28)$}& 46.53(4.21)& 32.51(2.77) \\
\hline
 \multirow{5}{*}{Caucasian} 
& baseline & 99.66&  97.60&  91.61  &81.01&  67.82&  55.16 \\
& UIR &  99.66(0.00)&  97.60(0.00)&  91.53(-0.08)&  80.80(-0.21)&  67.52(-0.30)&  54.80(-0.36) \\
& ARL &  99.67(0.01)&  97.53(-0.07)&  91.47(-0.14)&  81.11(0.10)&  67.94(0.12)&  55.15(-0.01) \\ 
& ARL+C &  99.67(0.01)&  97.59(-0.01)& 91.50(-0.11)& 81.15(0.14)&  68.17(0.35)&  55.48(0.32) \\ 
& ARL+C+G &99.67(0.01)&  97.56(-0.04)& 91.64(0.03)&  81.11(0.10)&  67.58(-0.24)& 54.94(-0.22) \\    \hline
 \multirow{5}{*}{Indian} 
& baseline & 99.00&  93.93&  83.63&  70.20&  55.73&  42.65 \\
& UIR &  99.06(0.06)&  94.39(0.46)&  84.46(0.83)&  71.22(1.02)&  56.78(1.05)&  43.60(0.95) \\
& ARL &  99.09(0.09)&  94.53(0.60)&  85.50(1.87)&  72.55(2.35)&  58.30(2.57)&  45.91(3.26) \\
& ARL+C &   \bm{$99.29(0.29)$}&  \bm{$95.41(1.48)$}&  \bm{$87.20(3.57)$}&  \bm{$74.71(4.51)$}&  \bm{$60.99(5.26)$}&  45.57(2.92) \\
& ARL+C+G & 99.17(0.17)&  95.19(1.26)&  86.93(3.30)&  74.53(4.33)&  60.76(5.03)&  \bm{$46.96(4.31)$} \\ \hline
\end{tabular}
\label{VGGFace2}
}
\end{table}

\begin{table}[!htbp]
\centering
\caption{Experiments on MS\_wo\_RFW \cite{RFW}. ``+C'' represents applying cosine penalty strategy, and ``+G'' means taking gender in to account for image selection.}
\scalebox{0.8}{
\begin{tabular}{c|c|cccccc}
\hline
\multirow{2}{*}{Ethnicity}& \multirow{2}{*}{Model}& \multicolumn{6}{c}{TPR@FPR} \\ \cline{3-8}
& & 1e-1 & 1e-2 & 1e-3 & 1e-4 & 1e-5 & 1e-6 \\ \hline
\multirow{6}{*}{African}
& baseline &  99.46&  97.60&  91.94&  83.28&  72.18&  61.93 \\
& UIR &  99.56(0.10)&  97.81(0.21)&  93.16(1.22)&  85.19(1.91)&  74.19(2.01)&  63.84(1.91) \\
& TCP &  99.59(0.13)&  97.85(0.25)&  92.69(0.75)&  84.04(0.76)&  72.52(0.34)&  61.53(-0.40) \\
& ARL &  99.60(0.14)&  90.03(0.43)&  93.71(1.77)&  85.90(2.62)&  75.92(3.74)&  65.22(3.29) \\ 
& ARL+C &   \bm{$99.65(0.19)$}&  \bm{$98.15(0.55)$}&  \bm{$94.01(2.07)$}&  \bm{$86.92(3.64)$}&  77.24(5.06)&  65.95(4.02) \\
& ARL+C+G & 99.60(0.14)&  \bm{$98.15(0.55)$}&  93.88(1.94)&  86.63(3.35)&  \bm{$77.40(5.22)$}&  \bm{$66.81(4.88)$} \\\hline
\multirow{6}{*}{Asian} 
&baseline &  99.60&  97.13&  90.73&  80.64&  67.41&  54.98 \\
& UIR &  99.67(0.07)&  97.29(0.16)&  91.09(0.36)&  80.77(0.13)&  67.15(-0.26)&  54.77(-0.21) \\
& TCP &  \bm{$99.71(0.11)$}&  97.39(0.26)&  91.05(0.32)&  80.18(-0.46)& 66.55(-0.86)&  53.21(-1.77) \\
& ARL &  99.64(0.04)&  97.56(0.43)&  91.68(0.95)&  81.72(1.08)&  69.40(1.99)&  56.12(1.14) \\
& ARL+C &   99.70(0.10)&  \bm{$97.58(0.45)$}&  91.24(0.51)&  80.75(0.11)&  67.67(0.26)&  55.05(0.07) \\
& ARL+C+G & 99.67(0.07)&  97.64(0.51)&  \bm{$92.12(1.39)$}&  \bm{$83.02(2.38)$}&  \bm{$70.95(3.54)$}&  \bm{$58.46(3.48)$} \\
\hline
\multirow{6}{*}{Caucasian} 
&baseline &  99.87&  99.27&  97.92  &95.11&  89.71&  82.44 \\
& UIR &  99.88(0.01)&  99.25(-0.02)&  97.74(-0.18)&  94.89(-0.22)&  89.29(-0.42)&  81.70(-0.74) \\
& TCP &  99.87(0.00)&  99.34(0.07)&   97.95(0.03)&   95.05(-0.05)&  89.48(-0.23)&  81.62(-0.82) \\
& ARL &  99.88(0.01)&  99.34(0.07)&   97.84(-0.08)&  95.09(-0.02)&  89.43(-0.28)&  81.81(-0.63) \\
& ARL+C &   99.88(0.01)&  99.29(0.02)&  97.99(0.07)  &95.08(-0.03)&  89.59(-0.12)&  81.39(-1.05) \\
& ARL+C+G & 99.90(0.03)&  99.36(0.09)&  90.00(0.08)  &95.04(-0.07)&  89.65(-0.06)&  81.44(-1.00) \\   \hline
\multirow{6}{*}{Indian} 
& baseline &  99.74&  98.40&  94.58&  87.81&  79.44&  69.81 \\
& UIR &  99.79(0.05)&  98.62(0.22)&  95.40(0.82)&  88.97(1.16)&  81.33(1.89)&  70.92(1.11) \\
& TCP &  \bm{$99.85(0.11)$}&  98.70(0.30)&  95.51(0.93)&  89.00(1.19)&  80.96(1.52)&  70.42(0.61) \\
& ARL &  99.76(0.02)&  98.68(0.28)&  95.44(0.86)&  89.40(1.59)&  81.42(1.98)&  71.28(1.47) \\ 
& ARL+C &   99.80(0.06)&  \bm{$98.80(0.40)$}&  \bm{$95.76(1.18)$}&  \bm{$89.97(2.16)$}&  82.14(2.70)&  71.86(2.05) \\
& ARL+C+G & 99.80(0.06)&  98.71(0.31)&  95.60(1.02)&  89.75(1.94)&  \bm{$82.26(2.74)$}&  \bm{$72.55(2.74)$} \\ \hline
\end{tabular}
\label{MS1M}
}
\end{table}

\begin{table}[h]
\centering
\caption{Median of normalized cosine similarity histograms on test dataset. ``P'' represents positive pairs, while ``N'' represents negative pairs.}
\scalebox{0.8}{
\begin{tabular}{c|cc|c|cc}
\hline
             & baseline& ARL  &           & baseline & ARL \\ \hline
African (P) &   0.584 & 0.574& Asian (P) &  0.570   & 0.563 \\
African (N) &   0.035 & 0.015& Asian (N) &  0.041   & 0.020 \\ \hline
Difference  &   0.549 & \bm{$0.559$}& Difference&  0.529   & \bm{$0.543$} \\ \hline
Caucasian (P)&   0.579 & 0.579& Indian (P)&  0.606   & 0.590 \\
Caucasian (N)&  -0.001 &-0.002& Indian (N)&  0.045   & 0.025 \\ \hline
Difference  &   0.580 & \bm{$0.582$}& Difference&  0.561   & \bm{$0.565$} \\ \hline
\end{tabular}
}
\label{Median}
\end{table}

Performance increment comes from more separable histograms of positive pairs and negative pairs. Take the experiment on MS\_wo\_RFW \cite{RFW} as an example. From Table \ref{Median}, we can clearly see the median difference of normalized cosine similarity histograms becomes larger.

\subsubsection{Purpose of $K$}
We introduce $K$ unlabeled images with a different ethnicity tag into the loss calculation. However, from the formulation of Arcface \cite{Arcface}, we know that these items affects the loss calculation marginally since cosine similarity for images from different ethnicity is close to 0. It is illustrated by the result shown in Table \ref{K}. We introduce this $K$ here to claim that we can make use of some id-related tag information in unlabeled dataset, such as ethnicity, gender.

\begin{table}[!htbp]
\centering
\caption{Comparison on with and without $K$}
\scalebox{0.8}{
\begin{tabular}{c|c|cccccc}
\hline
\multirow{2}{*}{Ethnicity}& \multirow{2}{*}{Model}& \multicolumn{6}{c}{TPR@FPR} \\ \cline{3-8}
& & 1e-1 & 1e-2 & 1e-3 & 1e-4 & 1e-5 & 1e-6 \\ \hline
 \multirow{2}{*}{African} 
& with $K$    &  97.97&  88.70&  73.56&  56.70&  41.45&  29.04 \\
& without $K$ &  98.24&  88.65&  73.48&  56.47&  40.49&  27.02 \\ \hline
 \multirow{2}{*}{Asian} 
& with $K$    &  97.38&  87.31&  71.40&  53.94&  38.09&  26.30 \\
& without $K$ &  97.46&  87.66&  71.47&  53.53&  37.55&  24.60 \\
\hline
 \multirow{2}{*}{Caucasian} 
& with $K$    & 99.46&  96.08&  87.90&  76.86&  64.24&  51.50 \\
& without $K$ & 99.41&  95.91&  87.77&  76.50&  62.99&  50.07 \\   \hline
 \multirow{2}{*}{Indian} 
& with $K$    & 98.85&  93.14&  83.06&  70.14&  57.05&  43.60 \\
& without $K$ & 99.00&  93.62&  83.59&  70.50&  57.31&  44.96 \\ \hline
\end{tabular}
\label{K}
}
\end{table}

\subsubsection{Make Full Use of Unlabeled Data}
According to the results in Table \ref{CASIA}, \ref{VGGFace2}, and \ref{MS1M}, we can find out that the model becomes fairer. It means that cosine penalty strategy takes advantage of unlabeled data better and further improve the performance on under-represented groups. It is because cosine penalty strategy requires the face recognition model to distinguish more identities and learn more from unlabeled images. Nevertheless, from comparison, the improvement comes from cosine penalty is relatively less significant than ARL.

Following RFW's protocol \cite{RFW}, we also evaluated on the provided difficult pairs. From the results in \ref{acc}, we can see that our method can increase the average accuracy (AVG) with nearly unchanged Caucasian accuracy. And the standard deviation (STD) shrinks, which means fairer.
\begin{table}[h]
\label{acc}
\caption{Verficition accuracy ($\%$) on difficult pairs in RFW \cite{RFW}. }
\centering
\scalebox{0.8}{
\begin{tabular}{c|cccc|c|c}
\hline
Model & African & Asian & Caucasian & Indian & AVG   & STD  \\\hline
baseline(CASIA-Webface \cite{CASIA}) & 82.85   & 82.68 & 91.52     & 85.50  & 85.64 & 4.13 \\
ARL+C(CASIA-Webface \cite{CASIA})    & 85.35   & 84.55 & 91.25     & 88.28  & 87.36 & 3.05 \\ \hline
baseline(VGGFace2 \cite{vggface2}) & 87.30   & 85.47 & 93.50     & 87.55  & 88.46 & 3.48 \\
ARL+C(VGGFace2 \cite{vggface2})    & 88.57   & 87.65 & 93.48     & 89.35  & 89.76 & 2.57 \\ \hline
baseline(MS\_wo\_RFW \cite{RFW}) & 92.03   & 91.05 & 97.18     & 93.78  & 93.51 & 2.69 \\
ARL+C(MS\_wo\_RFW \cite{RFW})    & 93.38   & 91.45 & 97.10     & 94.88  & 94.20 & 2.38 \\ \hline
\end{tabular}
}
\end{table}

Even though cosine penalty strategy shows its strength, we find that group Asian's performance drops with cosine penalty strategy when the threshold is strict in experiment on MS\_wo\_RFW \cite{RFW}. We manually check the negative pairs with high cosine similarity and we observe that nearly all of them are pairs of female faces. It demonstrates that our model is affected by gender imbalance.

\subsubsection{Study on Gender Balance Issue}
As we described above, MS\_wo\_RFW \cite{RFW} and RFW test dataset \cite{RFW} is bias with gender. Group Asian is the most balanced group in RFW test dataset and MS\_wo\_RFW \cite{RFW} is biased on gender.

With such an observation, we use a tool \cite{attr} to estimate the gender of unlabeled images. This tool gives out a score of masculine in the range of $[0, 1]$. Therefore, we label images with score lower than 0.3 as female, images with score larger than 0.7 as male, and images in the range of $[0.3, 0.7]$ as unknown. In order to be fair with respect to gender, we try to sample the same number of images for male and female. When we select unlabeled images, we first select from female sub-group. If we cannot select sufficient images, unknown images are supplemented. Then we finish selection with sampling male images. 

From the results in Table \ref{MS1M}, we can clearly see that the performance on group Asian increases when threshold. However, the balanced gender strategy doesn't affect experiments on CASIA-Webface \cite{CASIA} and VGGFace2 \cite{vggface2} that much. It is because these two labeled dataset is relatively balanced on gender.

\section{Conclusion}
By drawing ROC of baseline models, it's easy to observe that face recognition models trained on commonly used dataset have significant racial bias. Based on extensive experiments, our Asymmetric Rejection Loss can mitigate the racial bias in face recognition model due to introducing under-represented unlabeled images and a better optimization goal. Moreover, Cosine penalty strategy can further boost the performance on under-represented ethnicity group. In addition, our method doesn't require the unlabeled images having good cluster property that it will works well even if each subject in unlabeled dataset has only one image. Besides, face recognition models also suffer from gender imbalance, and a gender balanced selection on unlabeled images can help the model alleviate the gender bias.

\bibliography{ref.bib}

\begin{thebibliography}{10}

\bibitem{alvi2018turning}
Mohsan Alvi, Andrew Zisserman, and Christoffer Nell{\aa}ker.
\newblock Turning a blind eye: Explicit removal of biases and variation from
  deep neural network embeddings.
\newblock In {\em Proceedings of the European Conference on Computer Vision
  (ECCV)}, pages 0--0, 2018.

\bibitem{amini2019uncovering}
Alexander Amini, Ava Soleimany, Wilko Schwarting, Sangeeta Bhatia, and Daniela
  Rus.
\newblock Uncovering and mitigating algorithmic bias through learned latent
  structure.
\newblock 2019.

\bibitem{vggface2}
Qiong Cao, Li~Shen, Weidi Xie, Omkar~M Parkhi, and Andrew Zisserman.
\newblock Vggface2: A dataset for recognising faces across pose and age.
\newblock In {\em 2018 13th IEEE International Conference on Automatic Face \&
  Gesture Recognition (FG 2018)}, pages 67--74. IEEE, 2018.

\bibitem{chapelle2009semi}
Olivier Chapelle, Bernhard Scholkopf, and Alexander Zien.
\newblock Semi-supervised learning (chapelle, o. et al., eds.; 2006)[book
  reviews].
\newblock {\em IEEE Transactions on Neural Networks}, 20(3):542--542, 2009.

\bibitem{R3}
Ken Chen, Yichao Wu, Haoyu Qin, Ding Liang, Xuebo Liu, and Junjie Yan.
\newblock R3 adversarial network for cross model face recognition.
\newblock In {\em Proceedings of the IEEE Conference on Computer Vision and
  Pattern Recognition}, pages 9868--9876, 2019.

\bibitem{Arcface}
Jiankang Deng, Jia Guo, Niannan Xue, and Stefanos Zafeiriou.
\newblock Arcface: Additive angular margin loss for deep face recognition.
\newblock In {\em Proceedings of the IEEE Conference on Computer Vision and
  Pattern Recognition}, pages 4690--4699, 2019.

\bibitem{gao2017semi}
Yuan Gao, Jiayi Ma, and Alan~L Yuille.
\newblock Semi-supervised sparse representation based classification for face
  recognition with insufficient labeled samples.
\newblock {\em IEEE Transactions on Image Processing}, 26(5):2545--2560, 2017.

\bibitem{ms1m}
Yandong Guo, Lei Zhang, Yuxiao Hu, Xiaodong He, and Jianfeng Gao.
\newblock Ms-celeb-1m: A dataset and benchmark for large-scale face
  recognition.
\newblock In {\em European Conference on Computer Vision}, pages 87--102.
  Springer, 2016.

\bibitem{han2018face}
Chunrui Han, Shiguang Shan, Meina Kan, Shuzhe Wu, and Xilin Chen.
\newblock Face recognition with contrastive convolution.
\newblock In {\em Proceedings of the European Conference on Computer Vision
  (ECCV)}, pages 118--134, 2018.

\bibitem{resnet}
Kaiming He, Xiangyu Zhang, Shaoqing Ren, and Jian Sun.
\newblock Deep residual learning for image recognition.
\newblock In {\em Proceedings of the IEEE conference on computer vision and
  pattern recognition}, pages 770--778, 2016.

\bibitem{LFW}
Gary~B Huang, Marwan Mattar, Tamara Berg, and Eric Learned-Miller.
\newblock Labeled faces in the wild: A database forstudying face recognition in
  unconstrained environments.
\newblock 2008.

\bibitem{IJB-A}
Brendan~F Klare, Ben Klein, Emma Taborsky, Austin Blanton, Jordan Cheney,
  Kristen Allen, Patrick Grother, Alan Mah, and Anil~K Jain.
\newblock Pushing the frontiers of unconstrained face detection and
  recognition: Iarpa janus benchmark a.
\newblock In {\em Proceedings of the IEEE conference on computer vision and
  pattern recognition}, pages 1931--1939, 2015.

\bibitem{liu2017sphereface}
Weiyang Liu, Yandong Wen, Zhiding Yu, Ming Li, Bhiksha Raj, and Le~Song.
\newblock Sphereface: Deep hypersphere embedding for face recognition.
\newblock In {\em Proceedings of the IEEE conference on computer vision and
  pattern recognition}, pages 212--220, 2017.

\bibitem{liu2018transductive}
Yu~Liu, Guanglu Song, Jing Shao, Xiao Jin, and Xiaogang Wang.
\newblock Transductive centroid projection for semi-supervised large-scale
  recognition.
\newblock In {\em Proceedings of the European Conference on Computer Vision
  (ECCV)}, pages 70--86, 2018.

\bibitem{liu2018exploring}
Yu~Liu, Fangyin Wei, Jing Shao, Lu~Sheng, Junjie Yan, and Xiaogang Wang.
\newblock Exploring disentangled feature representation beyond face
  identification.
\newblock In {\em Proceedings of the IEEE Conference on Computer Vision and
  Pattern Recognition}, pages 2080--2089, 2018.

\bibitem{shufflenetv2}
Ningning Ma, Xiangyu Zhang, Hai-Tao Zheng, and Jian Sun.
\newblock Shufflenet v2: Practical guidelines for efficient cnn architecture
  design.
\newblock In {\em The European Conference on Computer Vision (ECCV)}, September
  2018.

\bibitem{mobilenetv2}
Mark Sandler, Andrew Howard, Menglong Zhu, Andrey Zhmoginov, and Liang-Chieh
  Chen.
\newblock Mobilenetv2: Inverted residuals and linear bottlenecks.
\newblock In {\em Proceedings of the IEEE Conference on Computer Vision and
  Pattern Recognition}, pages 4510--4520, 2018.

\bibitem{triplet_loss}
Florian Schroff, Dmitry Kalenichenko, and James Philbin.
\newblock Facenet: A unified embedding for face recognition and clustering.
\newblock In {\em The IEEE Conference on Computer Vision and Pattern
  Recognition (CVPR)}, June 2015.

\bibitem{contrastive_loss}
Yi~Sun, Yuheng Chen, Xiaogang Wang, and Xiaoou Tang.
\newblock Deep learning face representation by joint
  identification-verification.
\newblock In {\em Advances in neural information processing systems}, pages
  1988--1996, 2014.

\bibitem{efficientnet}
Mingxing Tan and Quoc~V Le.
\newblock Efficientnet: Rethinking model scaling for convolutional neural
  networks.
\newblock {\em arXiv preprint arXiv:1905.11946}, 2019.

\bibitem{devil}
Fei Wang, Liren Chen, Cheng Li, Shiyao Huang, Yanjie Chen, Chen Qian, and Chen
  Change~Loy.
\newblock The devil of face recognition is in the noise.
\newblock In {\em Proceedings of the European Conference on Computer Vision
  (ECCV)}, pages 765--780, 2018.

\bibitem{cosface}
Hao Wang, Yitong Wang, Zheng Zhou, Xing Ji, Dihong Gong, Jingchao Zhou, Zhifeng
  Li, and Wei Liu.
\newblock Cosface: Large margin cosine loss for deep face recognition.
\newblock In {\em Proceedings of the IEEE Conference on Computer Vision and
  Pattern Recognition}, pages 5265--5274, 2018.

\bibitem{RFW}
Mei Wang, Weihong Deng, Jiani Hu, Xunqiang Tao, and Yaohai Huang.
\newblock Racial faces in the wild: Reducing racial bias by information
  maximization adaptation network.
\newblock In {\em The IEEE International Conference on Computer Vision (ICCV)},
  October 2019.

\bibitem{center_loss}
Yandong Wen, Kaipeng Zhang, Zhifeng Li, and Yu~Qiao.
\newblock A discriminative feature learning approach for deep face recognition.
\newblock In {\em European conference on computer vision}, pages 499--515.
  Springer, 2016.

\bibitem{attr}
wondonghyeon.
\newblock Gender and race classification with face images.
\newblock \url{https://github.com/wondonghyeon/face-classification}.

\bibitem{xie2019self}
Qizhe Xie, Eduard Hovy, Minh-Thang Luong, and Quoc~V Le.
\newblock Self-training with noisy student improves imagenet classification.
\newblock {\em arXiv preprint arXiv:1911.04252}, 2019.

\bibitem{legonet}
Zhaohui Yang, Yunhe Wang, Chuanjian Liu, Hanting Chen, Chunjing Xu, Boxin Shi,
  Chao Xu, and Chang Xu.
\newblock Legonet: Efficient convolutional neural networks with lego filters.
\newblock In {\em International Conference on Machine Learning}, pages
  7005--7014, 2019.

\bibitem{CASIA}
Dong Yi, Zhen Lei, Shengcai Liao, and Stan~Z Li.
\newblock Learning face representation from scratch.
\newblock {\em arXiv preprint arXiv:1411.7923}, 2014.

\bibitem{UIR}
Haiming Yu, Yin Fan, Keyu Chen, He~Yan, Xiangju Lu, Junhui Liu, and Danming
  Xie.
\newblock Unknown identity rejection loss: Utilizing unlabeled data for face
  recognition.
\newblock In {\em Proceedings of the IEEE International Conference on Computer
  Vision Workshops}, pages 0--0, 2019.

\bibitem{zhan2018consensus}
Xiaohang Zhan, Ziwei Liu, Junjie Yan, Dahua Lin, and Chen Change~Loy.
\newblock Consensus-driven propagation in massive unlabeled data for face
  recognition.
\newblock In {\em Proceedings of the European Conference on Computer Vision
  (ECCV)}, pages 568--583, 2018.

\bibitem{zhao2011semi}
Xuran Zhao, Nicholas Evans, and Jean-Luc Dugelay.
\newblock Semi-supervised face recognition with lda self-training.
\newblock In {\em 2011 18th IEEE International Conference on Image Processing},
  pages 3041--3044. IEEE, 2011.

\bibitem{zhu2005semi}
Xiaojin~Jerry Zhu.
\newblock Semi-supervised learning literature survey.
\newblock Technical report, University of Wisconsin-Madison Department of
  Computer Sciences, 2005.

\end{thebibliography}
\bibliographystyle{plain}

\end{document}